%% file: main.tex
\documentclass[letterpaper, 10 pt, conference]{ieeeconf}  % Comment this line out if you need a4paper

\IEEEoverridecommandlockouts                          

\usepackage{graphics} % for pdf, bitmapped graphics files
\usepackage{epsfig} % for postscript graphics files
\usepackage{mathptmx} % assumes new font selection scheme installed
\usepackage{times} % assumes new font selection scheme installed
\usepackage{amsmath} % assumes amsmath package installed
\usepackage{amssymb}  % assumes amsmath package installed
\usepackage{microtype}
\usepackage{float}
\usepackage{stfloats}
\usepackage{url}

\usepackage{hyperref} % Xuhong Add
\hypersetup{
    colorlinks=true,
    linkcolor=black,
    filecolor=magenta,      
    urlcolor=black,
}

\usepackage[spaces,hyphens]{xurl}

\usepackage[noend]{algpseudocode}

\title{\LARGE \bf
TransWorldNG: Traffic Simulation via Foundation Model
}

\author{Ding Wang$^{1,\dagger}$, Xuhong Wang$^{1,\dagger}$, Liang Chen$^{1}$, Shengyue Yao$^{1}$, Ming Jing$^{2}$, Honghai Li$^{3}$,\\
Li Li$^{4}$, \textit{Fellow, IEEE}, Shiqiang Bao$^{2}$, Fei-Yue Wang$^{5}$, \textit{Fellow, IEEE}, and Yilun Lin$^{1,*}$, \textit{Member, IEEE}
\thanks{This work is supported by the Shanghai Artificial Intelligence Laboratory.}% <-this % stops a space
\thanks{$\dagger$ Equal contribution.}
\thanks{*Corresponding author: Yilun Lin (linyilun@pjlab.org.cn)}
\thanks{$^{1}$Ding Wang (wangding@pjlab.org.cn), Xuhong Wang (wangxuhong@pjlab.org.cn), Liang Chen (chenliang@pjlab.org.cn), Shengyue Yao (yaoshengyue@pjlab.org,cn), Yilun Lin (linyilun@pjlab.org.cn) are with Urban Computing Lab, Shanghai Artificial Intelligence Laboratory, Shanghai, China.}%
\thanks{$^{2}$Shiqiang Bao (baoshiqiang@51world.com.cn), Ming Jing (jingming@51world.com.cn) are with 51 World, Shanghai, China.}%
\thanks{$^{3}$Honghai Li (honghai\_1@126.com) is with the Research and Development Center of Transport Industry of Autonomous Driving Technology, RIOH High and Technology Group.}%
\thanks{$^{4}$Li Li (li-li@tsinghua.edu.cn) is with the Department of Automation, Tsinghua University, Beijing, China.}
\thanks{$^{5}$Fei-Yue Wang (feiyue.wang@ia.ac.cn) is with the Institute of Automation, Chinese Academy of Sciences, Beijing, China, and the Macau Institute of Systems Engineering, Macau University of Science and Technology, Macau, China.}
}

\begin{document}

\maketitle
\thispagestyle{empty}
\pagestyle{empty}

%%%%%%%%%%%%%%%%%%%%%%%%%%%%%%%%%%%%%%%%%%%%%%%%%%%%%%%%%%%%%%%%%%%%%%%%%%%%%%%%
\begin{abstract}
Traffic simulation is a crucial tool for transportation decision-making and policy development. However, achieving realistic simulations in the face of the high dimensionality and heterogeneity of traffic environments is a longstanding challenge. In this paper, we present TransWordNG, a traffic simulator that uses Data-driven algorithms and Graph Computing techniques to learn traffic dynamics from real data. The functionality and structure of TransWorldNG are introduced, which utilize a foundation model for transportation management and control. The results demonstrate that TransWorldNG can generate more realistic traffic patterns compared to traditional simulators. Additionally, TransWorldNG exhibits better scalability, as it shows linear growth in computation time as the scenario scale increases. To the best of our knowledge, this is the first traffic simulator that can automatically learn traffic patterns from real-world data and efficiently generate accurate and realistic traffic environments.
\end{abstract}

\input{contents/sec1.tex}
\input{contents/sec2.tex}
\input{contents/sec3.tex}

\input{contents/sec4.tex}
\input{contents/sec5.tex}
\input{contents/conclusion.tex}

\bibliographystyle{IEEEtran}

\bibliography{references}

\end{document}

%% file: contents/sec1.tex
\section{Introduction}
Modeling and simulating transportation systems realistically pose a challenge due to the high variability and diversity of traffic behaviors, as well as the spatial and temporal fluctuations that are difficult to model. Various traffic simulation models such as SUMO \cite{lopez_microscopic_2018}, MATSim \cite{andreas_introducing_2016}, AimSun \cite{aimsun_aimsun_2022}, VISSIM \cite{ptv_vissim_2012}, and others have been developed to simulate traffic systems with diverse scales. Although these models are useful, they still encounter limitations in realistically simulating the growing complexity and heterogeneity of urban transportation systems due to the restricted capability of the underlying parametric models and manually encoded rules \cite{yan_learning_2023}. To address this gap, advanced traffic simulation techniques are necessary that can generate more realistic traffic behaviors from real-world data \cite{fan_multi-agent_2021,li_simulation_2023}. This is critical for aiding traffic planners and policymakers in making well-informed decisions.

Traditional approaches often rely on physical dynamic models and implement data-driven approaches to learn parameters in the pre-defined models \cite{wang_mixture_2020}. However, such approaches may introduce oversimplifications and assumptions that curtail their accuracy and applicability \cite{barcelo_fundamentals_2010}. As a result, traditional models are suitable for specific tasks but not scalable or extensible, posing challenges in adapting to varying environments and managing large and complex data inputs. Furthermore, the intrinsic complexity of transportation systems, influenced by diverse agents and factors that affect traffic behavior, makes it a challenging task to realistically capture the temporal variability and complexity of traffic conditions. The dynamic and constantly evolving nature of transportation environments necessitates a flexible approach to simulating the traffic system that can quickly adapt to changes in the environment.

\begin{figure*}[ht]
\centering
\includegraphics[width=0.9\linewidth]{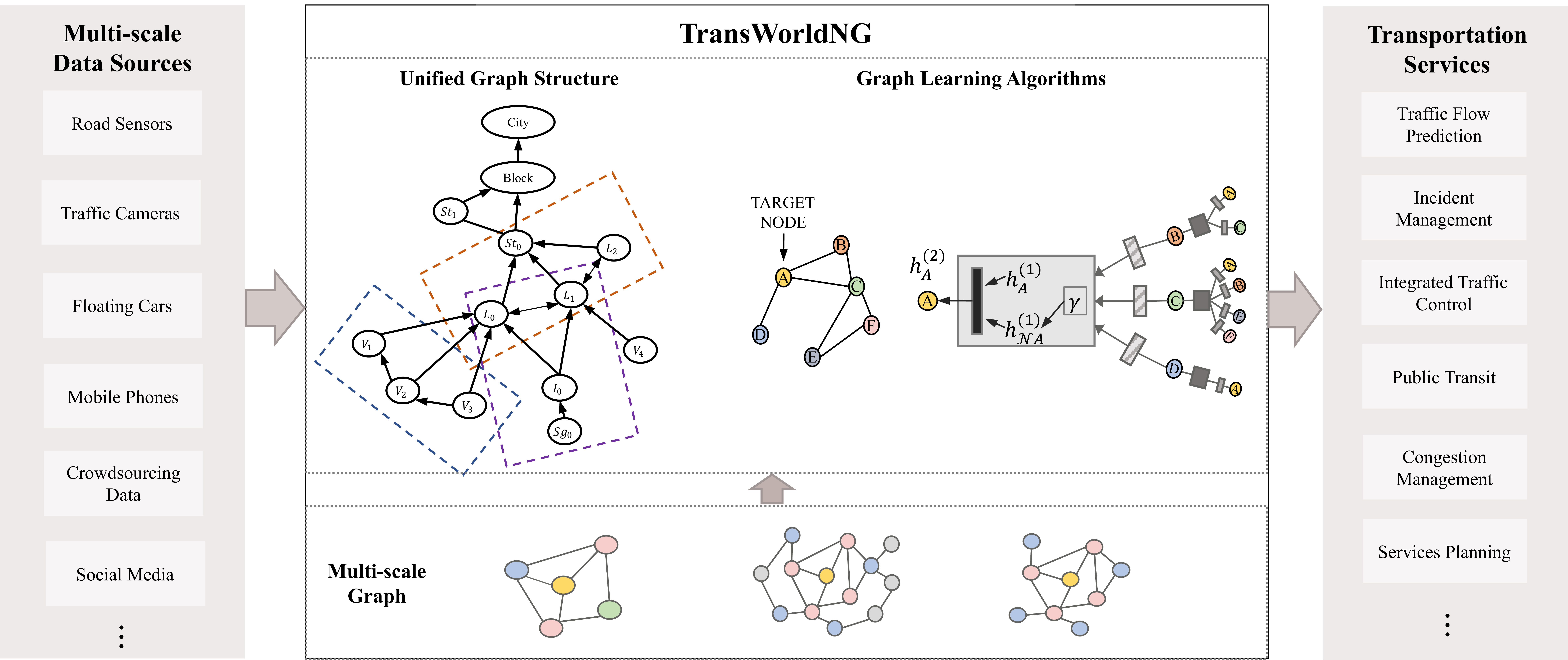}
\caption{The framework of TransWorldNG. TransWorldNG is built upon data-driven approaches, with the ability to handle multi-scale and multi-source data. This flexibility enables TransWorldNG to be used for a wide range of traffic-related tasks, making it a powerful tool for accurate and realistic simulations of urban transportation systems.}
\label{fig_foundation_model}
\end{figure*}

To solve these problems in traffic simulation, we have developed TransWorldNG (where NG denotes the new generation), which automatically generates simulation scenarios from multi-scale and high-dimensional data, the framework of TransWorldNG is shown in Fig. \ref{fig_foundation_model}. The first generation of TransWorld was initially developed by CAST Lab that uses Agent-based modeling (ABM) technology and object-oriented programming~\cite{wang_transportation_2017, wang_capturing_2018,li_simulation_2023}. Building on its framework, we have re-designed a data-driven traffic simulator that is empowered by the foundation model utilizing Data-driven algorithms and Graph Computing techniques to simulate intricate traffic systems~\cite{wang_building_2023}.

One of the key features of TransWorldNG is the utilization of graph structures and dynamic graph generation algorithms to model the intricate relationships and interactions among agents in the traffic system. This approach enhances previous ABM-based traffic simulation techniques by providing a more comprehensive and adaptable representation of the changing environment. Additionally, the use of graph structures and dynamic graph generation algorithms can enhance the scalability and efficiency of TransWorldNG by enabling parallel processing of the simulation and supporting the handling of large-scale data.

To overcome the limitations of traditional modeling approaches that rely on physical dynamics models, TransWorldNG adopts a data-driven approach with behavior models that are directly learned from real-world data. This approach provides a more direct and dependable representation of the real scenario. Furthermore, the graph structure of TransWorldNG allows for adaptive scaling, which amplifies its flexibility. Users can easily modify the nodes or edges in the graph structure to input multi-source data, with varying degrees of granularity.

This study presents the functionality and structure of TransWorldNG, the contributions of this paper are as follows:

\begin{itemize}
\item A unified graph-based structure has been proposed that permits a flexible representation of the varying traffic environments, facilitating TransWorldNG to adapt to the environment changes in real-time.
\item A data-driven traffic simulation framework has been introduced which can realistically and efficiently learn traffic dynamics from real-world data.
\item The underlying software designing principles, comprising of system structure, workflows, and interfaces for users, have been provided.
\end{itemize}

%% file: contents/sec2.tex
\section{Related Works}

\subsection{Multi-agent Traffic Modeling and Simulation}
Agent-based modeling is a widely used technique for modeling and simulating transportation systems, which involves simulating the interactions of a large number of agents in a system with different characteristics, behaviors, and interactions with other agents~\cite{macal_agent-based_2009,nguyen_overview_2021}. 
% ABM has been applied at different scales, including microscopic, mesoscopic, and macroscopic approaches, to simulate complex transportation systems~\cite{lin_diversity_2019}. 
The theoretical framework has developed over several decades, including game theory, control theory, graph theory, and complex network theory \cite{macal_agent-based_2009, lin_diversity_2019}. 
% ABM has been applied in many fields to simulate individual behaviors, interactions between agents, and emergent macro-level behaviors in complex systems. 
Transportation systems can be considered as multi-agent systems composed of different types of traffic participants, each with their own goals and behaviors, and their interactions affect the changes in the entire traffic system. Recent research on modeling and simulation of complex traffic systems is mostly based on multi-agent methods~\cite{nguyen_overview_2021,chen_review_2010}.

The modeling of multi-agent systems involves using mathematical models to describe the behavior of individual agents or the entire system in order to better understand traffic evolution and complex traffic phenomena~\cite{wang_mobility_2021}. In the field of traffic systems, models are typically categorized into three types based on their modeling scales: macroscopic, mesoscopic, and microscopic models~\cite{barcelo_fundamentals_2010,nguyen_overview_2021,andreas_introducing_2016,zhang_cityflow_2019}.

\subsection{Data-driven Traffic Modeling and Simulation}
The advancement of modeling complex transportation systems is expected to be driven by the availability of large-scale and multi-source data~\cite{kevan_can_2020}. Data-driven techniques in transportation modeling utilize machine learning, deep learning, and other algorithms to analyze large-scale and multi-source data and learn rules directly from the data to models. This is in contrast to knowledge-driven approaches, which rely on human-defined rules and models to develop transportation models~\cite{wu_hybrid_2018}. Urban big data can be used to assess the effects of different characteristics, such as road network topology and intersection shapes, on traffic flow in urban areas. Machine learning techniques, such as neural networks, support vector machines, and regression trees, can be trained using these data to anticipate traffic flow, speed, and congestion~\cite{avila_data-driven_2020}. This can provide valuable insights into the behavior of urban transportation systems and inform effective transportation planning and management strategies. Previous data-driven approaches in transportation research are mostly employed for single-task research, such as forecasting vehicle trajectories, predicting traffic congestion, route optimization, and so on \cite{van_oort_data_2015, kerner_introduction_2009, jia_data_2018}. These approaches have limitations in their ability to handle the complex interactions between multiple types of agents in a heterogeneous environment for large-scale systems. 

%% file: contents/sec3.tex
\section{Framework, System Structure, and Workflows of TransWorldNG}

\subsection{The Framework of TransWorldNG}
\begin{figure}[h]
\centering
\includegraphics[width=1\linewidth]{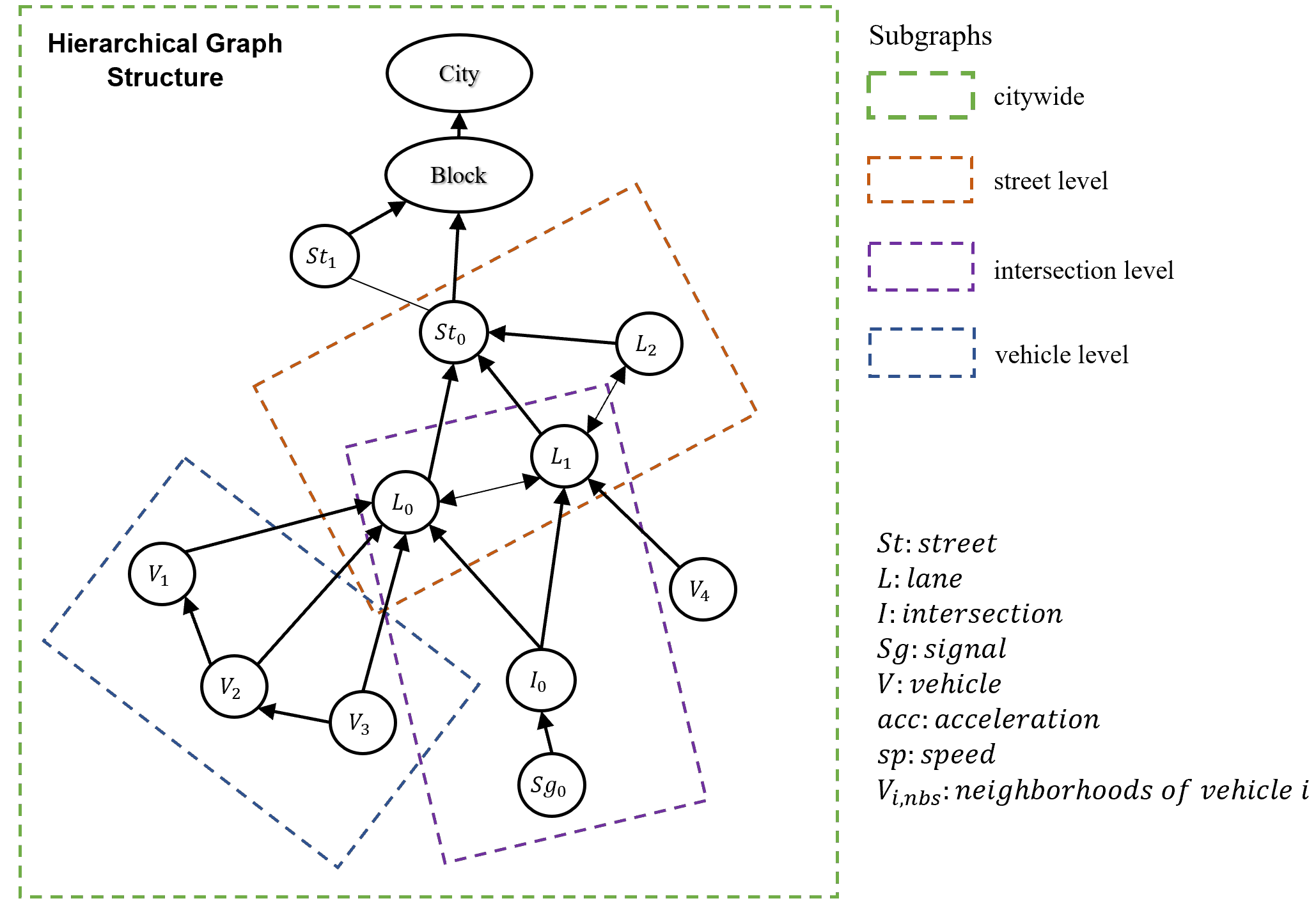}
\caption{Hierarchical graph structure of TransWorldNG. A hierarchical graph structure that consists of sub-graphs, with the lowest level often representing the finest granularity of simulated interactions. These sub-graphs are interconnected, allowing information to flow seamlessly through the different levels of the hierarchy.}
\label{fig_transcale}
\end{figure}

Transportation system modeling traditionally involves defining the behavior of agents and their interactions beforehand, which is time-consuming and error-prone when new agents or scenarios need to be added. A graph-based approach to transportation system modeling offers a more efficient and adaptable solution, as it allows for the representation of data and relationships between agents in a natural and straightforward way \cite{zhou_towards_2022, wu_comprehensive_2021}. By using a graph data structure, new data can be added to the system by introducing new nodes or edges to the graph, without the need to hard-code specific behaviors or rules.
Fig. \ref{fig_transcale} illustrates the topology of the hierarchical graph of the dynamic traffic system at various scales, from the vehicle level presenting the dynamics of individual vehicles to the intersection level showing traffic signal control strategies and traffic conditions at bottlenecks, to the street, block, and city levels showing strategies and policies that have impacts on a larger scale. This multi-scale approach allows TransWorldNG NG to provide a comprehensive view of the traffic system and enable transportation planners to make informed decisions based on the simulation results.

\subsubsection{Representation of transportation system via heterogeneous dynamic graphs} 
\begin{figure}[ht]
\centering
\includegraphics[width=1\linewidth]{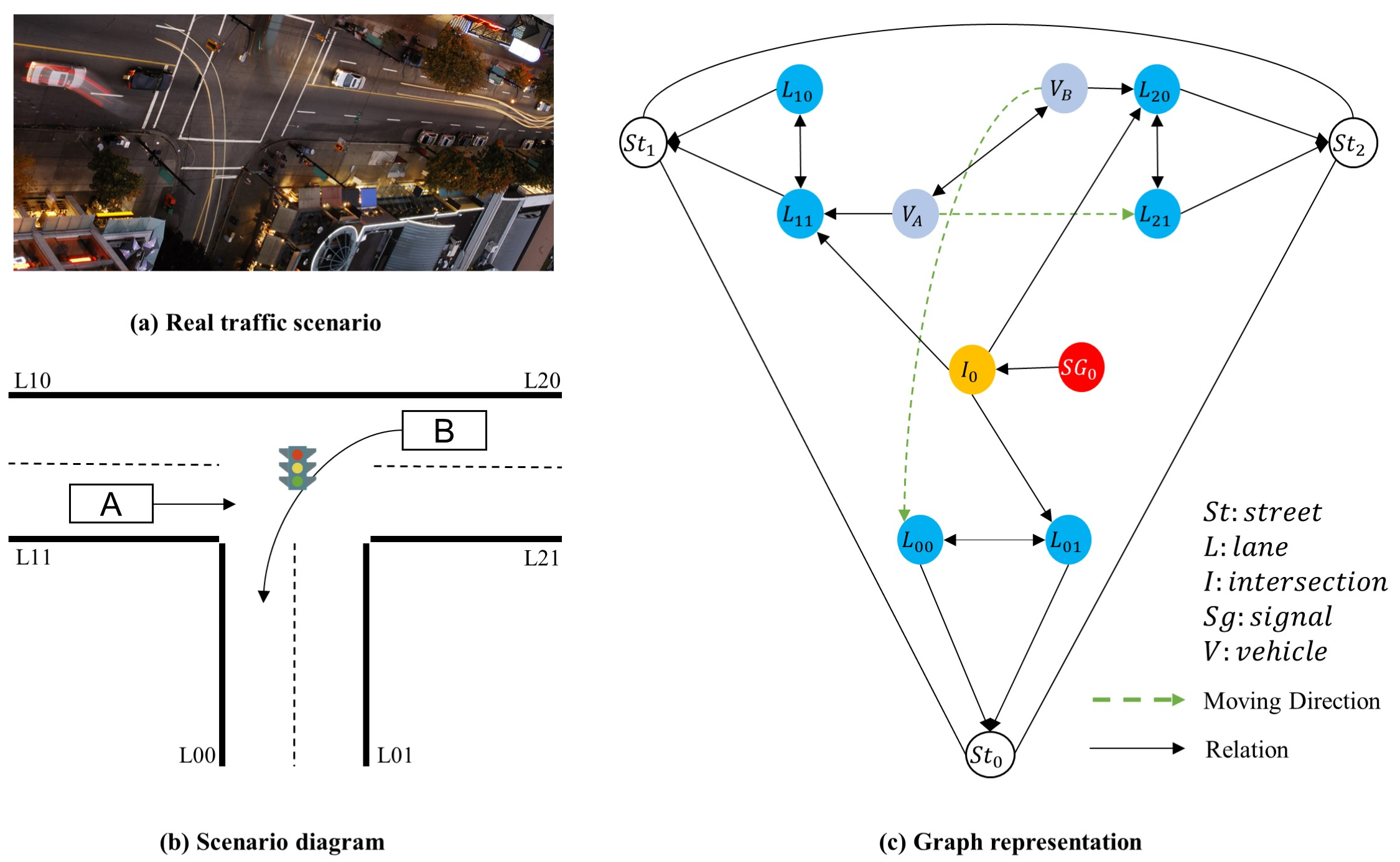}
\caption{Graph representation of a traffic scenario involving two cars, Car A moving straight from left to right and Car B turning left at a signalized intersection. (a) A picture of the real traffic scenario; (b) An abstract representation of the traffic scenario; (c) Graph representation of the traffic scenario. }
\label{fig_graph-represent-exp}
\end{figure}

TransWorldNG uses a unified graph data structure to represent traffic systems, this makes it flexible to changes in the environment, as it would allow for easy updates and modifications. New data can be added to the graph by introducing new nodes or edges, without the need to hard-code specific behaviors or rules. This flexibility and adaptability make it easier to model and simulate large and complex transportation systems. Fig. \ref{fig_graph-represent-exp} illustrates an example of a traffic scenario represented as a graph, showing how the relationships and interactions in a transportation system can be represented using a graph.

Mathematically, we can define the traffic system as a dynamic heterogeneous graph, $G_{n}(V_{n}, E_{n}, O_{n}, R_{n})$. The graph consists of vertices ($V_{n}$) that represent agents and edges ($E_{n}$) that define the relationships between those agents. Each agent ($v_i $) is associated with a node type by a unique mapping:

\begin{equation}
\phi: V_n \to O_n, o_i \in O_n  \\
\end{equation}

($\phi: V_n \to O_n$). Similarly, each edge ($e_i$) is directed and associated with an edge type:

\begin{equation}
\psi: E_n \to R_n, r_i \in R_n, e_i =(u_i,v_i)  \\
\end{equation}

The attributes of agents can be represented as node features on the graph. For instance, a vehicle agent might have attributes such as position, speed, and acceleration, which can be saved as node features. Assuming nodes and edges have feature dimensions $D^v_{o_i}$and  $D^e_{r_i}$ respectively, features can be represented as:

\begin{equation}
F^v_n \in \mathcal{R}^{|V_n| \times D^v_{o_i}},F^e_n \in \mathcal{R}^{|E_n| \times D^e_{r_i}} \\
\end{equation}

\subsubsection{Dynamic Graph Learning model to simulate traffic behavior and relationships} 
TransWorldNG can learn from the data and generate simulation scenarios without relying on pre-defined models or assumptions. 
% TransWorldNG uses graph learning models to simulate traffic behavior and relationships \cite{velickovic_graph_2017,hu_heterogeneous_2020}. 
The use of a data-driven and model-free approach allows TransWorldNG to discover new patterns and relationships in the data that may not have been previously known or considered. This can lead to insights and solutions that were not possible with traditional modeling approaches.

To simulate complex traffic behavior and relationships, the Heterogeneous Graph Transformer (HGT) model can be used to model heterogeneous graphs in transportation systems \cite{hu_heterogeneous_2020}. The HGT model is a powerful graph neural network that can handle the heterogeneity of graphs by utilizing specific representations for different types of nodes and edges. It uses a multi-head attention mechanism to aggregate information from neighbors of different node types, and a graph transformer layer to update the node embeddings, for details refer to ~\cite{hu_heterogeneous_2020}.

We denote the output of the $l$-th HGT layer as $H^{(l)}$, $H^{(l)}[v]$ is the node representation of node $v$ at the $l$-th HGT layer. By stacking $L$ layers, the node representation for the whole graph can be represented as $H^{(L)}$. Since the traffic network is time-varying, we consider the evolution of the traffic system as a conditional graph translation process that consists of a sequence of static graphs:

\begin{equation}
T: G_0 \rightarrow G_1 \dots \rightarrow G_{n} \\
\end{equation}

Given the dynamic heterogeneous graph, $G(V, E, O, R)$ shows the state of the traffic simulation system, with input node features, denoted as $H^{(l-1)[v]}$. The output of the $l$-th HGT layer for the target node $v$ is denoted as $H^{(l)}[v]$. To aggregate information to the target node from all its neighbors, it can be achieved by updating the vector $\tilde{H}^{(l)}[v]$:

\begin{equation}
\tilde{H}^{(l)}[v] = \sum_{\forall(u)\in N(v)}(Attention(u,e,v)\cdot Message(u,e,v)) \\
\end{equation}

The $l$-th HGT layer’s output $H^{(l)}[v]$ for the target node $v$ is equal to:

\begin{equation}
H^{(l)}[v] = A\text{-}Linear_{\phi(v)}(\theta\tilde{H}^{(l)}[v])+H^{(l-1)}[v] \\
\end{equation}

The MSE loss is employed to measure the difference between the predicted values and the true values. In practice, the MSE loss can be optimized using various optimization algorithms such as Stochastic Gradient Descent (SGD), Adam, or RMSProp to minimize the difference between predicted and true values~\cite{kingma_adam_2014,ruder_overview_2016}.

\subsection{System Structure}
\begin{figure}[ht]
\centering
\includegraphics[width=1\linewidth]{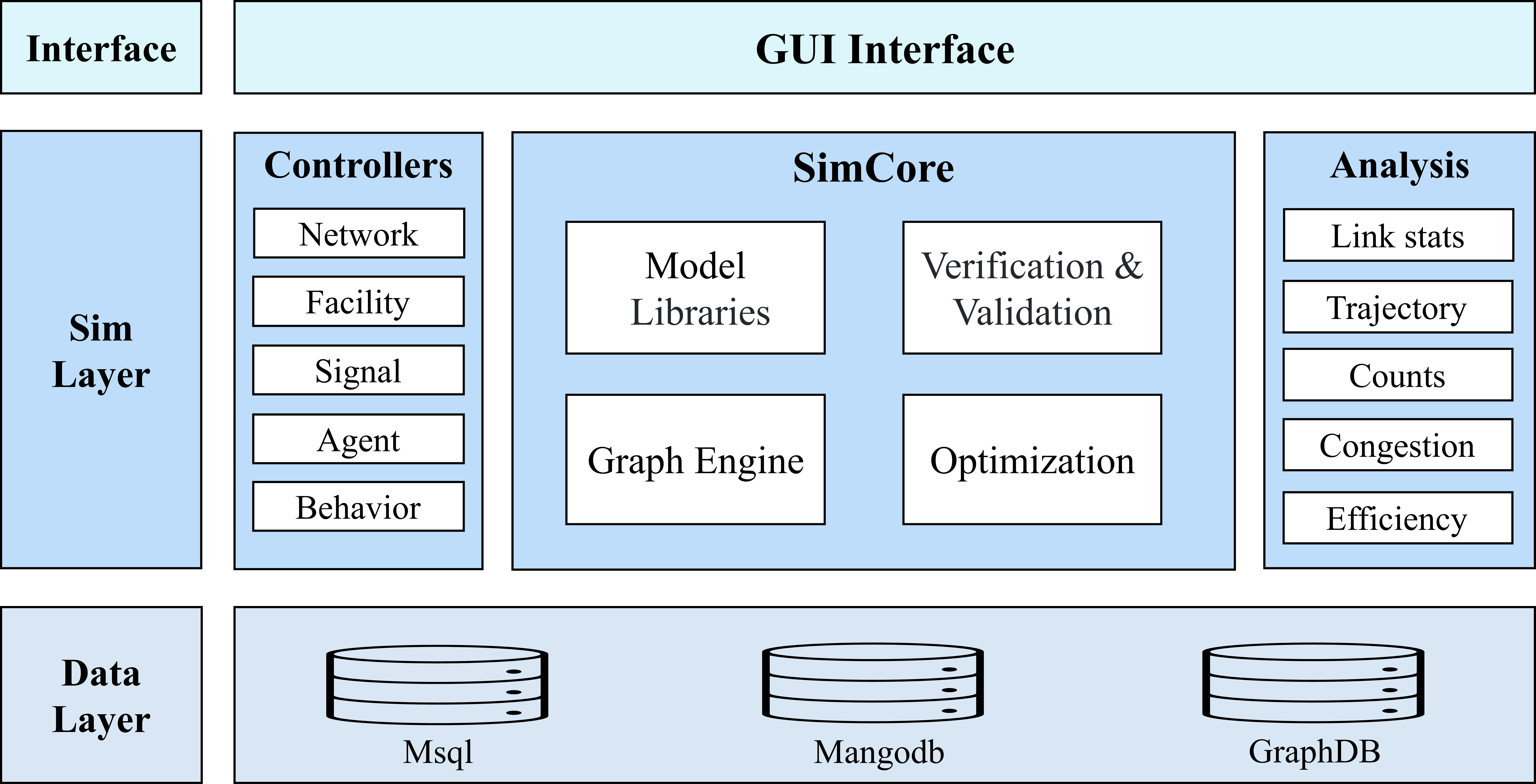}
\caption{Overall System Architecture of TransWorldNG. This figure illustrates the key components and their relationships. }
\label{fig_system-architecture}
\end{figure}

The overall system architecture of TransWorldNG is shown in Fig. \ref{fig_system-architecture}. The system supports data inputs from different sources, including sensors, GPS devices, and other connected devices. These data inputs are processed and transformed into a graph data structure, which is then fed into the simulation core in the simulation layer. Using mathematical models and algorithms, the simulation core simulates traffic flow, predicts congestion, and optimizes the transport network based on different traffic scenarios. The software layers can be divided into three categories: data layer, simulation layer, and interface layer.

\textbf{Data Layer:} The data layer includes both graph and non-graph data. The non-graph data is stored in a relational database like MySQL and Mangodb, while the graph data structure is stored in a graph database. This allows for the efficient handling of different types of data in a complementary manner.

\textbf{Simulation Layer:} The simulation layer includes the simulation core, which consists of the simulation core, controllers, and analysis modules. 
\begin{itemize} 
\item \textit{SimCore:} The simulation core consists of the model libraries, graph engine, optimization module, and verification and validation processes. The model libraries provide a range of models for simulating and analyzing the transportation network data, while the graph engine provides algorithms for processing and analyzing the graph data. The optimization module uses algorithms to find the optimal parameters for the models, and the verification and validation processes ensure that the data and results are accurate and reliable. 
\item \textit{Controller:} The controller module is responsible for controlling network dynamics, traffic signals, and agent behaviors, and uses the simulation core to simulate different scenarios.
\item \textit{Analysis:} The analysis module provides insights into the transportation network's performance by processing and analyzing simulation results, such as link statistics, trajectory analysis, traffic counts, congestion analysis, efficiency measures, and more.
\end{itemize}

\begin{figure}[ht]
\centering
\includegraphics[width=1\linewidth]{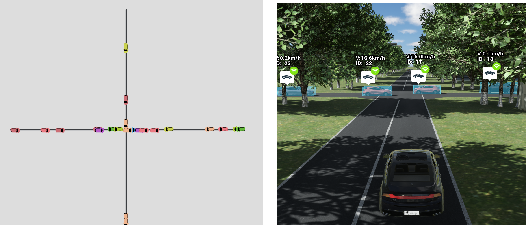}
\caption{GUI of TransWorldNG. The figure shows the graphical user interface of TransWorldNG, which is used to interact with the traffic simulator. The GUI is designed to be user-friendly and intuitive. The main window of the GUI displays a 3D visualization of the simulated traffic environment. }
\label{fig_gui}
\end{figure}

\textbf{Interface Layer:} The interface layer includes the GUI interface that displays simulation results to the user, shown in Fig \ref{fig_gui}. The GUI interface provides visualizations and graphs to help the user understand and interpret the simulation results.

\subsection{Workflow}

\begin{figure*}[ht]
\centering
\includegraphics[width=0.8\linewidth]{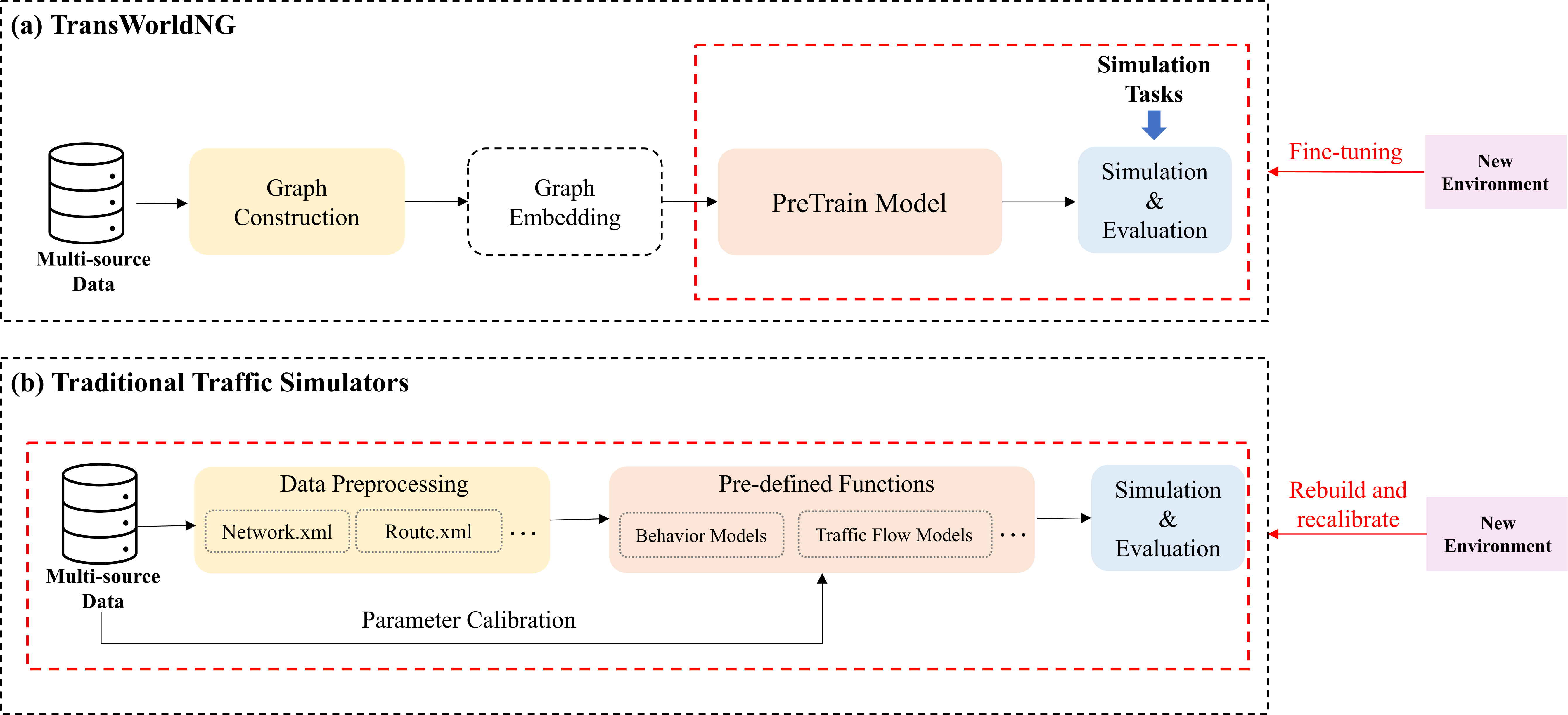}
\caption{Comparison of traffic simulation workflows: (a) TransWorldNG, which uses graph construction and embedding techniques to obtain a pre-trained model for different simulation tasks, and (b) Traditional traffic simulators that require building and calibrating pre-defined behavior models. When the environment changes to new states, TransWorldNG can quickly adapt by fine-tuning the pre-trained model with new data, while traditional simulators need to start from scratch and repeat the entire simulation process, which is highlighted in the red dotted box.}
\label{fig_workflow}
\end{figure*}

TransWorldNG  is designed to be intuitive and user-friendly. 
%The software provides a workflow that allows users to easily load graph-structured data, create and modify transportation networks, specify agent types and attributes, and modify the parameters of the graph learning model, such as batch size, training epoch, and loss functions. 
The simulation core generates traffic patterns based on the input data and parameters specified by the user. These traffic patterns can be visualized in real-time or exported for further analysis. The workflow of TransWorldNG compared to traditional simulation models can be found in Fig. \ref{fig_workflow}. The key modules in TransWorldNG are the following:

\begin{itemize}
\item \textit {Graph Construction:} TransWorldNG constructs a heterogeneous graph representation of the traffic environment from real-world traffic data. Nodes represent individual agents, while edges represent the relationships and interactions between agents.
\item \textit {Graph Embedding:} The graph is embedded into a high-dimensional space using a heterogeneous graph transformer model.
\item \textit {Pre-training:} The graph transformer model is pre-trained on a large dataset of real-world traffic scenarios, enabling it to learn the patterns and relationships.
\item \textit {Simulation:} The pre-trained is then used to generate simulations of new traffic scenarios. The system can make dynamic adjustments during the simulation to model changes in the traffic environment.
\item \textit {Evaluation:} The simulations generated by TransWorldNG can be evaluated based on various metrics, such as accuracy and efficiency, which can help to improve the system for better performance.
\end{itemize}

%% file: contents/sec5.tex
\section{CASE STUDY}
\begin{figure}[h]
\centering
\includegraphics[width=1\linewidth]{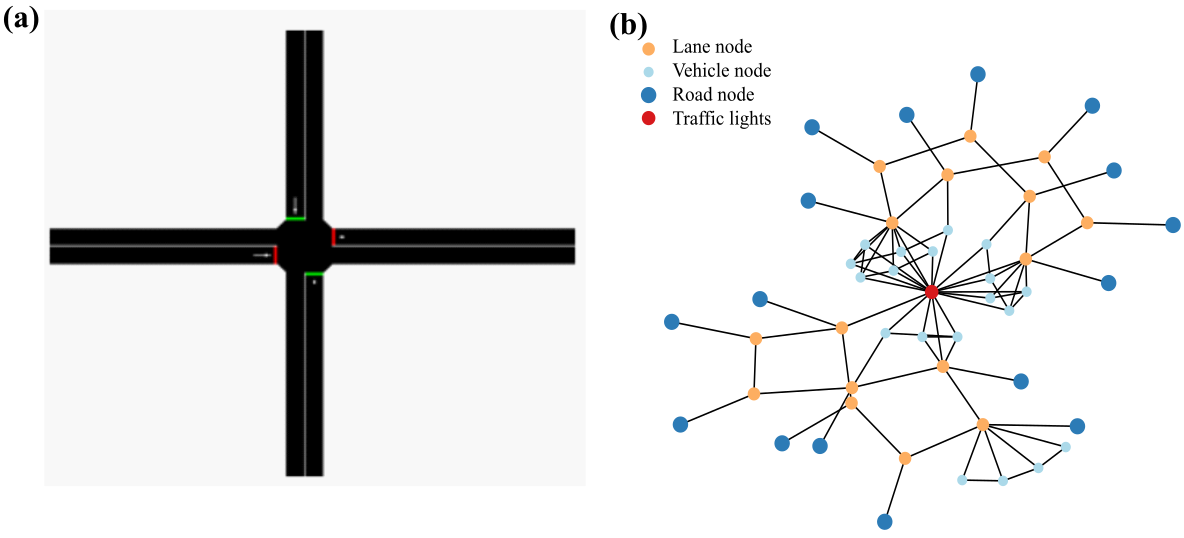}
\caption{The case study scenario: (a) Traffic network of the simulated environment and (b) corresponding graph representation of the traffic system at one time step. The graph representation captures the structure and connections of the traffic system. }
%Subplot (c) and (d) compare the features of vehicle speed and acceleration extracted by the TransWorldNG model and SUMO, respectively, highlighting the ability of the TransWorldNG model to generate similar results compared to the input data.}
\label{fig_feature_accurancy}
\end{figure}

This section aims to demonstrate the capabilities and advantages of TransWorldNG compared to existing traffic simulators. A case study is conducted to compare TransWorldNG with SUMO \cite{lopez_microscopic_2018}, a widely used traffic simulator. A 4-way signalized intersection is simulated using both TransWorldNG and SUMO. Fig. \ref{fig_feature_accurancy} (a) shows the 4-way signalized intersection, which is a classic example scenario in SUMO. The road network has 8 roads and 16 lanes, and there are 768 vehicles running in this network. These vehicles all start from the left direction and have three default routes they can take: going down to the right road, turning left to the north road, or turning right to the south road. The scenario also has one traffic light located at the central intersection.

% \begin{figure}[ht]
% \centering
% \includegraphics[width=1\linewidth]{images/simone.png}
% \caption{Screenshot. }
% \label{fig_Screenshot}
% \end{figure}

%A 4-way signalized intersection is simulated using both TransWorldNG and SUMO. The simulation aims to demonstrate the ability of TransWorldNG to accurately model the car-following model, lane-changing behavior, and signal control logic at the micro-scale.

\subsection{Data-Driven Traffic Behavior Learning with TransWorldNG}

\begin{figure}[h]
\centering
\includegraphics[width=0.9\linewidth]{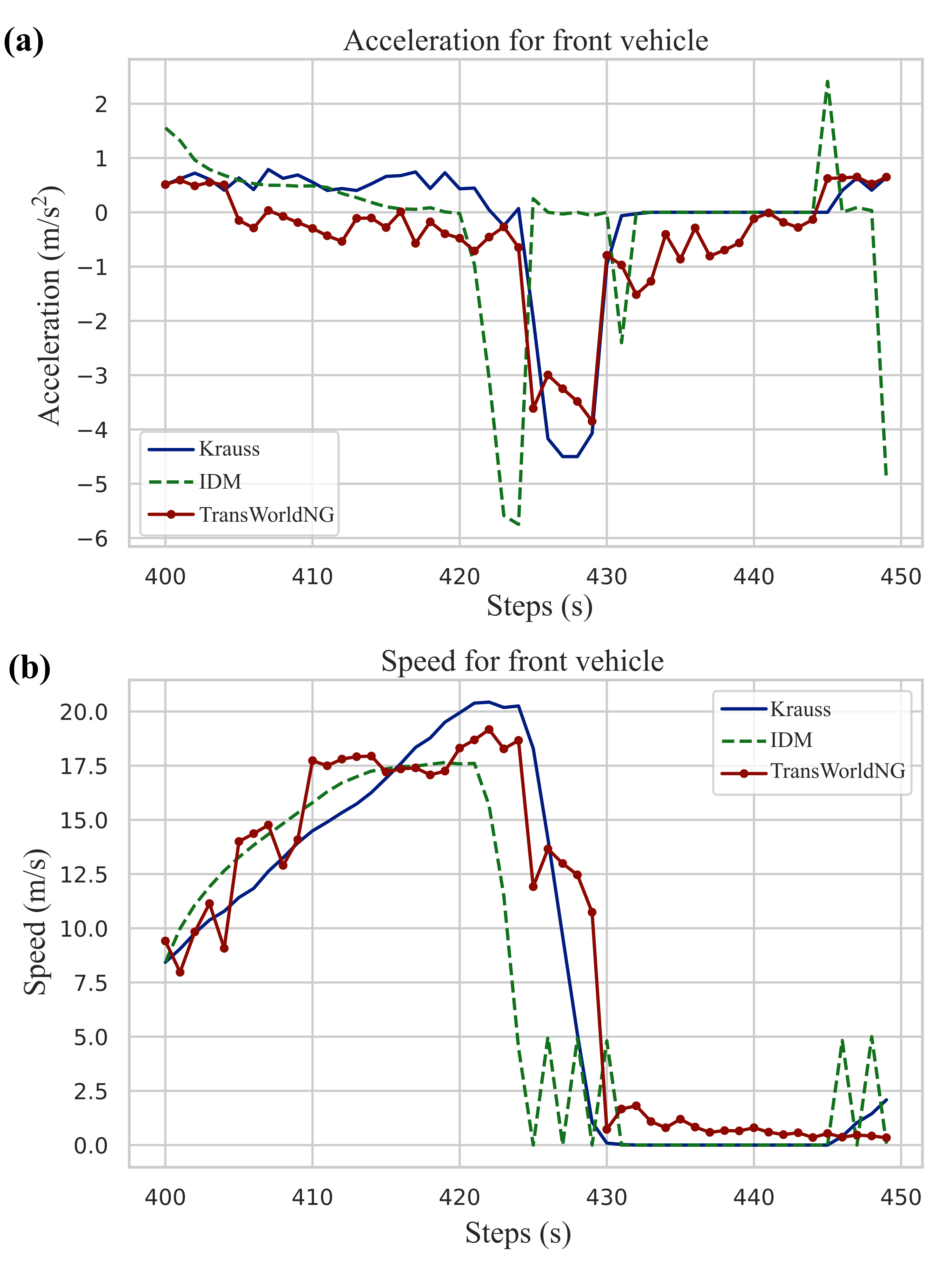}
\caption{Comparison of the car-following model generated by TransWorldNG, IDM, and the Krauss mode, which is the default car-following model of SUMO. Subplots (a) and (b) present the vehicle acceleration and speed for the front car, as predicted by the three models, respectively, showing the ability of the TransWorldNG model to learn car following behavior from data and can generate similar patterns compared to those well-known models.}
\label{fig_cf_behavior}
\end{figure}

We investigated the ability of TransWorldNG to learn car-following behaviors from data. To evaluate its performance, we compared the car-following behavior generated by TransWorldNG to the Intelligent Driver Model (IDM) and the Krauss model. The Krauss model is the default car-following model of the SUMO traffic simulation software. Fig. \ref{fig_cf_behavior} presents the comparison of vehicle acceleration and speed for the front car, as predicted by the three models. 

\begin{figure}[h]
\centering
\includegraphics[width=0.9\linewidth]{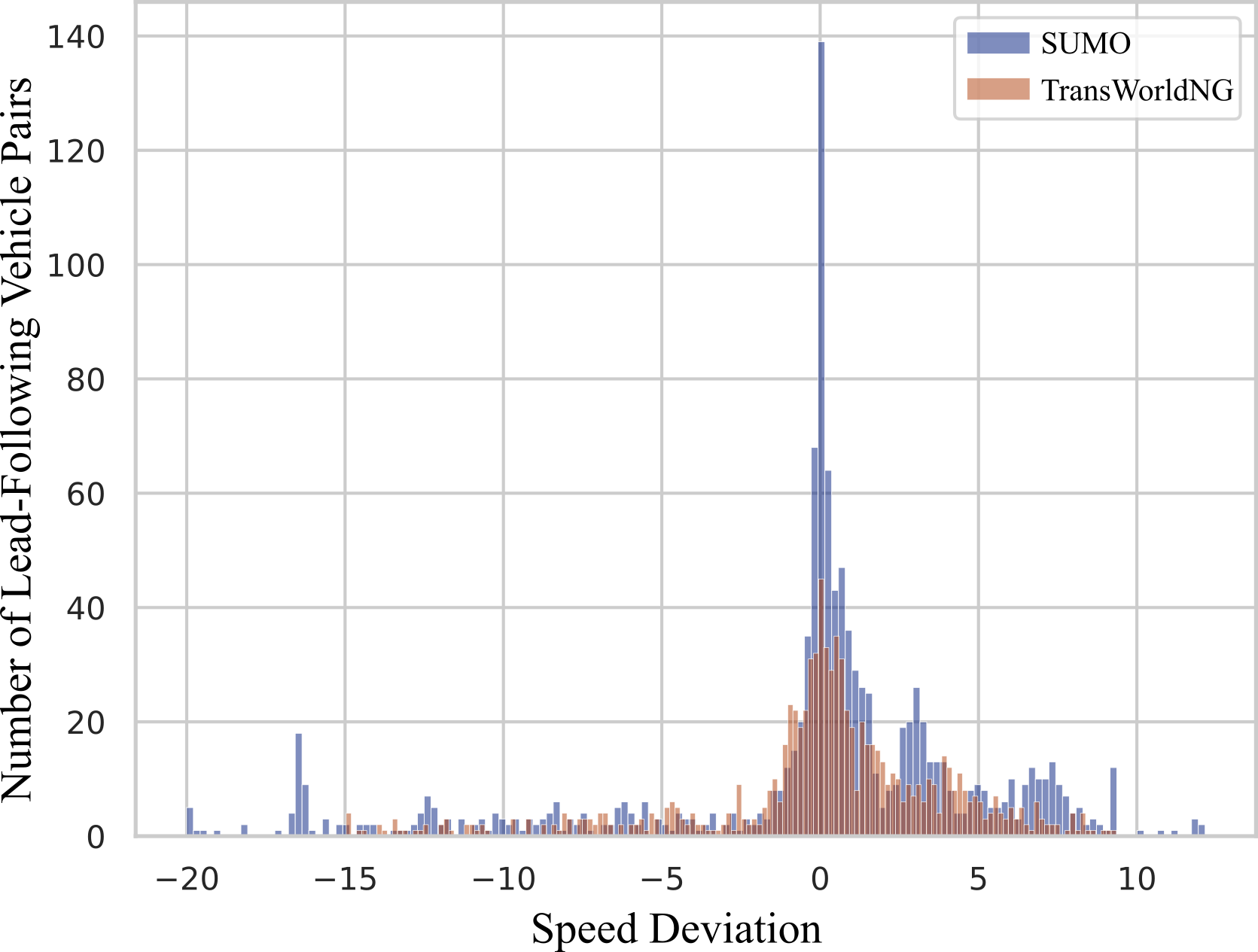}
\caption{Histogram of the distribution of speed deviation in the car following behavior. The speed deviation is defined as the speed difference between the lead and follower vehicles. A speed deviation of zero indicates that the front and follower vehicles are traveling at a relatively consistent speed.}
\label{fig_speed_dev}
\end{figure}

Fig.~\ref{fig_speed_dev} presents a histogram of the frequency of speed deviations observed during a simulation, showing the performance of the two simulation environments in terms of speed control and accuracy. A narrower distribution with a smaller spread and a peak closer to zero typically indicates better speed control and accuracy in the simulation environment. Both histograms in the figure show similar distributions, indicating that TransWorldNG performs car-following behavior as well as the classic models. This suggests that the automatically generated car-following behavior in TransWorldNG is effective.

\subsection{Impact of Data Collection Interval on Model Performance}
Since TransWorldNG is a data-driven approach, to understand the trade-off between prediction accuracy and data collection frequency in TransWorldNG, we conducted experiments with different data collection intervals (5 and 10 steps) and compared the results with SUMO. The findings reveal interesting insights into the relationship between data collection interval and prediction accuracy. As expected, with shorter data collection intervals (e.g., 5 steps), the TransWorldNG model can capture more frequent updates in traffic dynamics, resulting in higher prediction accuracy. However, as the data collection interval increases (e.g., 10 steps), the prediction accuracy decreases, indicating that the model's ability to capture real-time changes in traffic dynamics is reduced. These findings highlight the importance of data collection frequency in the TransWorldNG model and emphasize the need for careful consideration of data collection intervals to achieve optimal prediction accuracy.

\begin{figure}[h]
\centering
\includegraphics[width=0.95\linewidth]{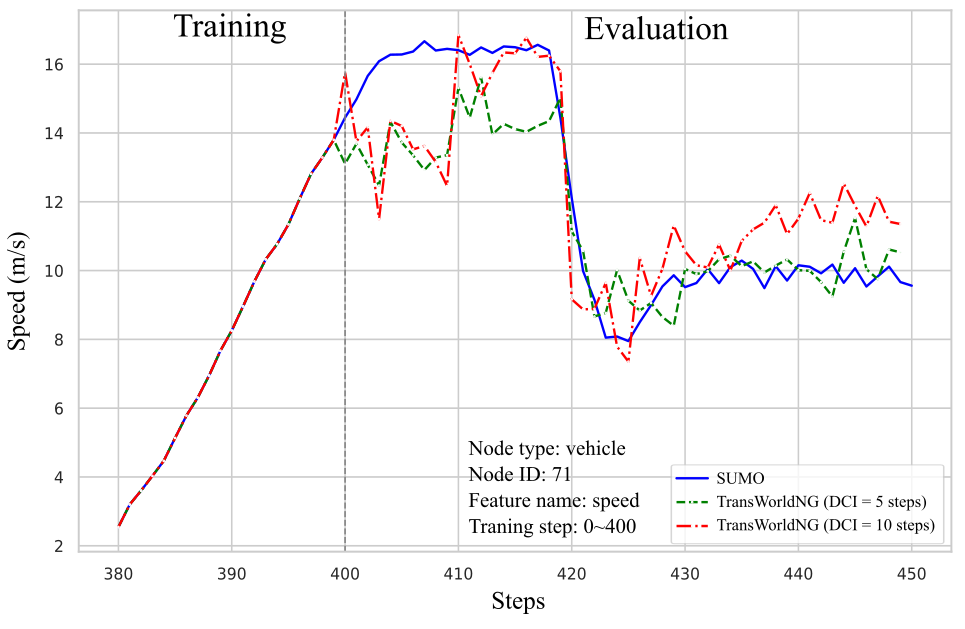}
\caption{Impact of Data Collection Interval (DCI) on Model Performance of TransWorldNG. This result shows the comparison of predicted vehicle speed between SUMO (as a reference) and TransWorldNG with data collection intervals of 5 and 10 steps, respectively. The result provides insights into the trade-off between prediction accuracy and data collection frequency in the TransWorldNG model.}
\label{fig_data_collection_interval}
\end{figure}

\subsection{Assessing the Computational Performance of TransWorldNG}
Evaluating the computational performance of TransWorldNG is an important aspect of assessing the system's efficiency and scalability for large-scale traffic simulations. Simulation time and the number of agents are typically inversely proportional, meaning that as the number of agents increases, the simulation time will also increase. One way to evaluate the computational performance of TransWorldNG is to measure the percentage increase in runtime as the percentage increase in the number of agents.

\begin{figure}[h]
\centering
\includegraphics[width=0.95\linewidth]{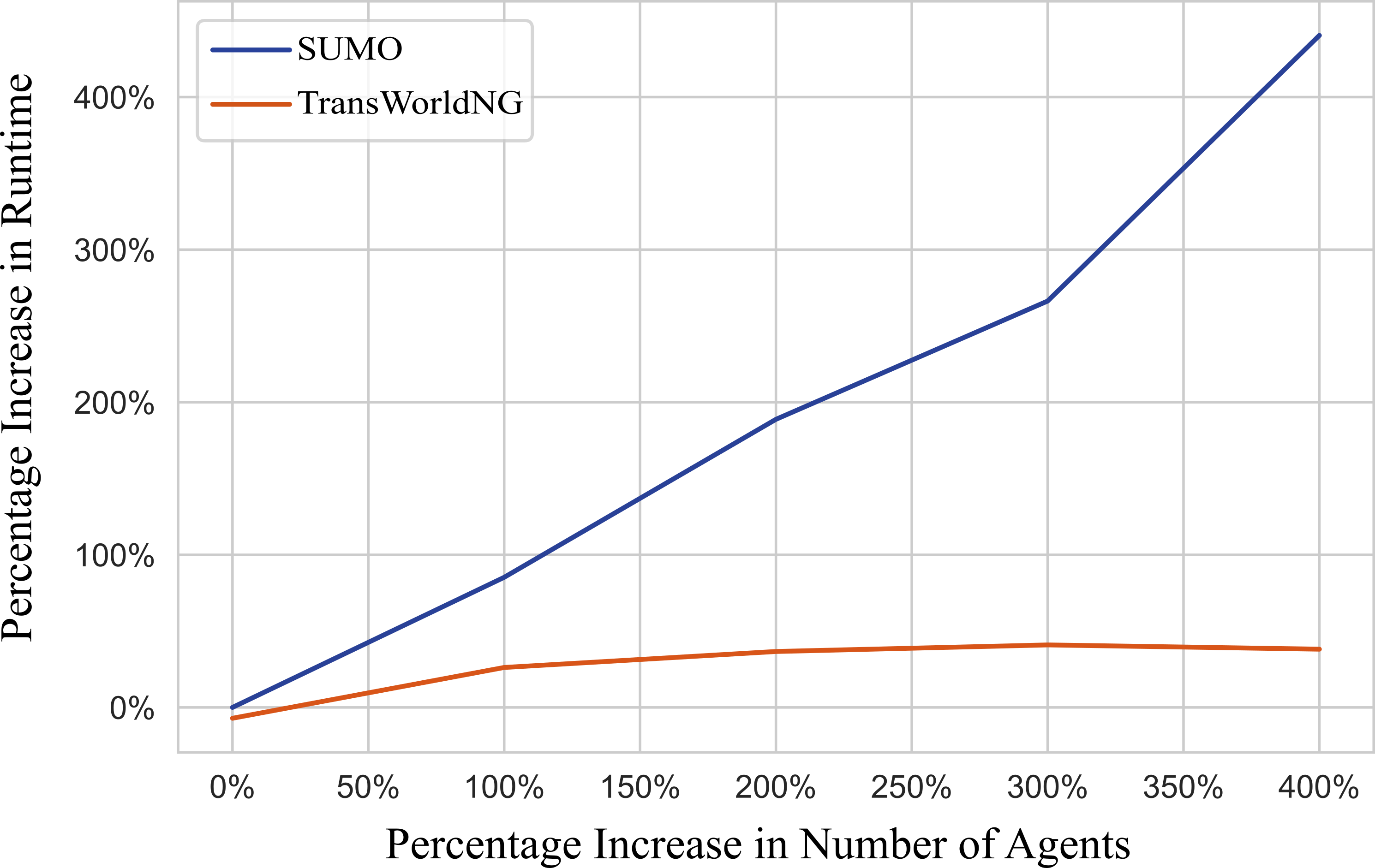}
\caption{Illustration of the computational efficiency of TransWorldNG compared to SUMO.}
\label{fig_computation_complex}
\end{figure}

Fig. \ref{fig_computation_complex} compares the percentage increase in simulation calculation time between TransWorldNG and SUMO as the system scale increases. The results demonstrate that as the system scale increases, the percentage increase in simulation calculation time grows substantially more slowly for TransWorldNG than for SUMO. This indicates that the proposed framework can dramatically improve the computing efficiency of large-scale traffic simulation. These results highlight the benefits of using TransWorldNG as a framework for large-scale traffic simulation.

One of the key reasons for TransWorldNG's good performance is its use of a graph structure and pre-trained models. The use of a graph structure enables parallel processing of traffic data, which can significantly reduce simulation calculation time. Additionally, the pre-trained models used in TransWorldNG can help reduce the amount of computation required during simulation, as the models have already learned many of the underlying patterns and relationships in the traffic data. Another factor that contributes to TransWorldNG's superior performance is its model-free approach. Unlike SUMO, which relies on pre-defined models of traffic behavior, TransWorldNG is able to adapt to different traffic scenarios and levels of abstraction without the need for extensive model development and calibration. This allows for more efficient and flexible simulation of complex traffic scenarios.

%% file: contents/conclusion.tex
\section{Conclusion}
% Modeling and simulation of complex transportation systems are critical for effective transportation management and control. Traditional simulations aim to replicate or approximate existing systems, while TransWorldNG takes a different approach by using real data to "grow" live traffic processes and generate alternative versions of real traffic activities. Unlike existing traffic simulations that focus only on direct traffic-related activities, TransWorldNG can include a wider range of factors that can influence traffic processes, such as weather, legal, and social factors. The unified graph structure and flexible adaptive scaling of TransWorldNG make it highly scalable and versatile, bridging the gap between virtual simulations and practical applications.

This study introduced the simulation framework and system structure of TransWorldNG, which utilize a traffic foundation model with data-driven automatic modeling capabilities to resolve the issues of limited structural complexity and high computation complexity of traditional simulators. The graph structure and data-driven method permit dynamic adjustments during simulation to reflect real-time changes in the urban system environment, allowing for the insertion of new data and expert knowledge for real-time mapping of the simulation system to the actual city. TransWorldNG can facilitate event-driven causal analysis of urban phenomena and can combine multi-field data to provide a simulation test platform for integrated decision-making. Future directions for TransWorldNG could include the integration of emerging technologies such as Mobility as a Service (MaaS) and AI-driven simulation technologies \cite{kevan_can_2020}, as well as the development of more robust functionality using the framework presented in this study.

While TransWorldNG offers many advantages for traffic simulation, there are also some potential challenges that should be explored in future research. One potential challenge is the need for high-quality data that accurately represent real-world traffic patterns and behaviors. TransWorldNG relies on large amounts of data to generate its simulations, so the accuracy and quality of this data can significantly impact the reliability and usefulness of the simulations. Additionally, the collection, processing, and storage of such large amounts of data can also be a challenge \cite{yu_swdpm_2023}. In addition, while TransWorldNG is designed to be highly scalable and flexible, it may still face challenges in terms of computational resources and processing power. Running large-scale simulations can require significant computing resources, potential solutions include using cloud computing, distributed computing, and parallel processing, which need to be studied in future research. Furthermore, traffic simulation often involves predicting traffic flow over extended time periods.  The potential use of large language models (LLMs), such as GPT, to generate a wider range of realistic scenarios may improve the accuracy and effectiveness of traffic simulations~\cite{wang_what_2023}.

\section*{Data Availability}
The SUMO simulation platform and data is publicly available at
\url{https://www.eclipse.org/sumo}.
The simulation data of the 4-way intersection scenario is available from SUMO at
\url{https://github.com/eclipse/sumo/blob/main/docs/web/docs/Tutorials}.

\section*{Code availability}
The code of TransWorldNG was implemented in Python using the deep learning framework of PyTorch. Code, trained models, and scripts reproducing the experiments of this paper are available at \url{https://github.com/PJSAC/TransWorldNG}.